\documentclass{article}
\usepackage{spconf,amsmath,amssymb,graphicx,hyperref}
\usepackage{cite}
\usepackage{algorithm,algpseudocode}
\usepackage{enumitem}
\usepackage{array,multirow,hhline}
\usepackage[svgnames,x11names]{xcolor}

 \newcommand{\alna}[1]{\begin{alignat}{3}&#1&\end{alignat}}
\def \diag {{\rm diag}}

\title{Adaptive Graph Coarsening for Efficient GNN Training}

\name{Rostyslav Olshevskyi, Madeline Navarro, and Santiago Segarra
\thanks{This work was partially supported by the NSF under award CCF-2340481. 
Research was sponsored by the Army Research Office and was accomplished under Grant Number W911NF-17-S-0002. The views and conclusions contained in this document are those of the authors and should  not be interpreted as representing the official policies, either expressed or implied, of the Army Research Office or the U.S. Army or the U.S. Government. The U.S. Government is authorized to reproduce and distribute reprints for Government purposes notwithstanding any copyright notation herein. 
    Emails:  
    \{\href{mailto:ro22@rice.edu}{ro22}, 
    \href{mailto:nav@rice.edu}{nav}, 
    \href{mailto:segarra@rice.edu}{segarra}\}@rice.edu }
}
\address{Rice University, Houston, TX, USA}

\begin{document}
\input{mysymbol.sty}
\ninept
\maketitle
\begin{abstract}
We propose an \textit{adaptive graph coarsening method} to jointly learn graph neural network (GNN) parameters and merge nodes via $K$-means clustering during training.
As real-world graphs grow larger, processing them directly becomes increasingly challenging and sometimes infeasible.
Tailoring algorithms to large-scale data may sacrifice performance, so we instead consider graph reduction to decrease the amount of data used during training.
In particular, we propose a method to simultaneously train a GNN and coarsen its graph by partitioning nodes via $K$-means clustering based on their embeddings.
Unlike past graph coarsening works, our approach allows us to merge nodes during training.
Not only does this preclude coarsening as a preprocessing step, but our node clusters can adapt to the learning task instead of relying solely on graph connectivity and features.
Thus, our method is amenable to scenarios that are challenging for other methods, such as heterophilic data.
We validate our approach on both homophilic and heterophilic node classification datasets.
We further visualize relationships between node embeddings and their corresponding clusters to illustrate that our coarsened graph adapts to the learning task during training.
\end{abstract}

\section{Introduction}
\label{sec:intro}

Learning on graph-structured data continues to show success in many applications, even as tasks become increasingly complex~\cite{wu2021ComprehensiveSurveyGraph}. 
However, the exponential growth of such data renders their scalable implementation computationally challenging~\cite{hashemi2024comprehensivesurveygraph,ma2025AccelerationalgorithmsGNNs}. 
Some approaches simplify graph-based learning algorithmically, which include distributing training or exploiting convergence properties of growing graphs~\cite{zhu2024SimplifyingDistributedNeural,cervino2023LearningbyTransference}.
However, these works still require processing the full-sized graph during training and may sacrifice performance for their modifications.

Recent data-centric perspectives instead aim to reduce the amount of data for training graph-based models~\cite{hashemi2024comprehensivesurveygraph,xu2024SurveyGraphCondensation,yin2025GraphCondensationFoundations,xu2025LearningReduceScale}.
Graph sparsification techniques inevitably discard data by employing subsets of nodes or edges~\cite{liu2022SamplingMethodsEfficient}.
Moreover, stochastic subsampling approaches may require many training iterations to extract necessary information~\cite{ma2025AccelerationalgorithmsGNNs}.
An alternative is to consolidate structure in the full graph, which can improve scalability while encoding relevant characteristics if the graph is efficiently represented~\cite{gao2025GraphCondensationSurvey}.
For example, graph condensation works design smaller, synthetic graphs to replace the original in training~\cite{gao2024Graphcondensationinductive,liu2024Graphdistillationeigenbasis,liu2024TinyGraphJointFeature}.
However, there is no direct, interpretable mapping between nodes in the full and condensed graph to investigate how the original nodes contributed to learning.

A classical approach to graph reduction is to coarsen the structure by merging nodes with similar characteristics.
Because clustering nodes is a well-established task, graph coarsening and its connection to graph neural networks (GNNs) have enjoyed considerable empirical and theoretical analysis~\cite{huang2021ScalingGraphNeural,joly2024Graphcoarseningmessagepassing,joly2025Taxonomyreductionmatrices}.
This typically involves a two-phase approach, which first combines partitions of nodes into supernodes then trains a GNN on the coarsened graph~\cite{dickens2024GraphCoarseningConvolution,xia2025GraphCoarseningSupervised}.
Many methods merge nodes to preserve structural properties that are unrelated to the downstream task~\cite{loukas2019GraphReductionSpectral,joly2024Graphcoarseningmessagepassing}.
Moreover, most works are designed for graph convolutional networks (GCNs) and may struggle with heterophilic data~\cite{dickens2024GraphCoarseningConvolution,xia2025GraphCoarseningSupervised}.
Hence, graph coarsening was eschewed in favor of condensation, as synthetic graphs can be learned jointly with GNN parameters~\cite{gao2025GraphCondensationSurvey}, despite challenges with interpretability and efficiency~\cite{hashemi2024comprehensivesurveygraph}.

We thus present a graph coarsening method to jointly train a GNN and coarsen its graph with an adaptive $K$-means approach.
In particular, we apply alternating minimization to update clusters of node embeddings if they become sufficiently different during training, which is analogous to solving a bilevel optimization problem common to graph distillation~\cite{gao2025GraphCondensationSurvey}.
Thus, unlike previous works on graph coarsening for GNNs, we do not require a two-phase process with a clustering algorithm specific to the learning task.
Moreover, as supernodes simply encourage similarity among embeddings, our approach is model- and task-agnostic, adaptable to settings that are challenging for other graph coarsening methods such as heterophilic graph data~\cite{zheng2024graphneuralnetworks}.
Finally, our coarsening approach allows us to visualize node clusters to verify that embeddings are grouped as expected.
We summarize our contributions below.
\begin{itemize}[left= 4pt .. 12pt, noitemsep]
    \item[1.] We propose a graph coarsening approach to accelerate GNN training that alternates between $K$-means clustering of node embeddings and optimizing model parameters.
    \item[2.] We introduce an adaptive algorithm that updates node clusters when embeddings grow sufficiently different, thereby avoiding repeated expensive clustering steps during training.
    \item[3.] We demonstrate our approach in benchmark datasets, highlighting the performance–efficiency trade-off across different coarsening levels.
\end{itemize}

\section{Problem formulation}\label{sec:problem}

Without loss of generality, we exemplify our approach for node-level predictions.
We consider a graph $\ccalG = (\ccalV,\ccalE)$ of $N$ nodes $\ccalV$ and edges $\ccalE \subseteq \ccalV \times \ccalV$, where the edge $(i,j) \in \ccalE$ if and only if nodes $i$ and $j$ are connected.
The graph structure can be conveniently encoded in its adjacency matrix $\bbA \in \reals^{N \times N}$, where $A_{ij} \neq 0$ if and only if $(i,j) \in \ccalE$, with nonzero entries representing edge weights.
We observe a set of $M$ graph signals per node, collected in the feature matrix $\bbX \in \reals^{N\times M}$.
Each node is assigned a label $\bby = [\bby_{\rm \scriptscriptstyle train}^\top, \bby_{\rm \scriptscriptstyle test}^\top]^\top \in \ccalY^N$, of which we only observe a subset $\bby_{\rm \scriptscriptstyle train}$, and our goal is to recover the remaining unseen labels $\bby_{\rm \scriptscriptstyle test}$.

Furthermore, we aim to find a partition of the nodes $\{ \ccalI_k \}_{k=1}^K$ where $\ccalI_j \cap \ccalI_k = \varnothing$ for any $j\neq k$ and $\bigcup_{k=1}^K \ccalI_k = \{1,2,\dots, N\}$.
Each of the $K$ clusters corresponds to a supernode of a coarsened graph $\ccalG' = (\ccalV',\ccalE')$ such that $|\ccalV'| = K$, which we construct by the reduction matrix $\bbP \in \{ 0,1 \}^{N \times K}$ such that $\bbP \bbone = \bbone$ and
\alna{
    P_{ik} 
    &~:=~&
    \begin{Bmatrix}
        1, & i \in \ccalI_k \\
        0, & {\rm otherwise}
    \end{Bmatrix},
\label{eq:reduce}}
where each column $\bbP_{:,k}$ indicates which nodes belongs to the $k$-th cluster, and $\bbC := \diag(\bbP^\top\bbone)$ is a diagonal matrix containing the number of nodes in each cluster. 
Then, we construct the adjacency matrix $\bbA' := \bbP^\top \bbA \bbP$ of the coarsened graph $\ccalG'$, which also has features $\bbX' := \bbC^{-1}\bbP^\top \bbX$~\cite{loukas2019GraphReductionSpectral}.

Given the coarsened graph $\ccalG'$, we aim to learn the parameters $\bbTheta$ of a GNN $f(\cdot,\cdot;\bbTheta):\reals^{N\times M} \times \reals^{N\times N} \rightarrow \reals^{N \times H}$ such that its output $\bbZ := f(\bbX,\bbA;\bbTheta)$ allows us to predict the labels $\hby = g(\bbZ)$, where $g:\reals^{N\times H} \rightarrow \ccalY^N$ corresponds to a classifier given the embeddings $\bbZ$.
Since we assume an architecture $f$ that is agnostic to graph size, where $\bbTheta$ merely transforms node features $\bbX$, we may apply an analogous GNN $f'(\cdot,\cdot;\bbTheta):\reals^{K\times M} \times \reals^{K \times K} \rightarrow \reals^{K\times H}$ that corresponds to the coarsened graph $\ccalG'$ while sharing parameters with $f(\cdot,\cdot;\bbTheta)$.
Then, we propose to solve the following problem
\begin{subequations}\label{eq:problem}
\alna{
    &\min_{\bbTheta}&~
    \ccalL \Big( 
        \bby_{\rm\scriptscriptstyle train},~\,
        \bbP^*_{\bbTheta} f'(\bbX', \bbA'; \bbTheta) 
    \Big)
&\label{eq:upper_problem}\\&
    &{\rm s.t.}&~
    \bbA' \!= \bbP^{*\top}_{\bbTheta} \bbA \bbP^*_{\bbTheta}, \,
    \bbX' \!= (\bbC_{\bbTheta}^{*})^{-1} \bbP^{*\top}_{\bbTheta} \bbX, \,
    \bbC_{\bbTheta}^* = \diag(\bbP_{\bbTheta}^{*\top}\bbone),
&\nonumber\\&
    &&~
    \bbP^*_{\bbTheta} \in 
        \argmin_{\bbP \in \ccalP} 
        \left\| (\bbP\bbC^{-1}\bbP^\top - \bbI) f( \bbX, \bbA; \bbTheta ) \right\|_F^2
\label{eq:lower_problem}}
\end{subequations}
where $\ccalL$ denotes a loss function to be minimized, such as cross entropy for classification error or mean squared error for regression, and $\ccalP$ denotes the set of valid reduction matrices~\cite{joly2025Taxonomyreductionmatrices}.
Observe that~\eqref{eq:upper_problem} optimizes GNN performance on the \textit{reduced} graph $\ccalG'$, seeking a coarsening such that the GNN output for each supernode improves performance for its corresponding nodes in the original graph.
The lower-level problem~\eqref{eq:lower_problem} returns the mapping from the original to the coarsened graph, which aims to preserve predictions between each node and its assigned supernode.
Thus, in addition to reducing training complexity, our graph coarsening implicitly encourages partitions of nodes with similar yet accurate predictions, which are obtained not from the original graph data but rather the task of interest via the upper-level problem~\eqref{eq:upper_problem}.
Indeed, the success of graph reduction relies on assuming an underlying simplicity in the learning task, and ours directly imposes this assumption at the GNN output.
This is particularly advantageous for node classification, where predictions belong to a finite, discrete set of targets $\ccalY$.

While our choice of architecture for $f$ is flexible, some common choices include the seminal GCN~\cite{kipf2017semisupervisedClassificationGraph}, which yields an $L$-layer output obtained by applying
\alna{
    \bbZ^{(\ell+1)}
    &~=~&
    \sigma\left(
        \tbA \bbZ^{(\ell)} \bbTheta^{(\ell)}
    \right)
\label{eq:gcn}}
at the $\ell$-th layer, where $\tbA := \tbD^{-\frac{1}{2}} (\bbA + \bbI) \tbD^{-\frac{1}{2}}$ denotes the normalized adjacency matrix with self loops for $\tbD := \diag((\bbA + \bbI)\bbone)$ and $\sigma$ denotes an elementwise nonlinearity.
However, as the low-pass GCN in~\eqref{eq:gcn} is highly effective on homophilic graph data but suffers in heterophilic settings, we may instead consider a multi-hop adaptation that learns a bank of polynomial graph filters~\cite{ruiz2021graph}, where the $\ell$-th layer of a filter-bank GCN (FBGCN) is
\alna{
    \bbZ^{(\ell + 1)}
    &~=~&
    \sigma\left(
        \sum_{r=0}^{R-1}
        \bbA^r \bbZ^{(\ell)} \bbTheta^{(\ell)}_r
    \right),
\label{eq:fbgcn}}
where a separate weight matrix $\bbTheta_r^{(\ell)}$ is learned at each $r$-hop neighborhood.
Thus, a single layer of a FBGCN increases the radius of the architecture without requiring several layers, mitigating the problem of oversmoothing~\cite{zheng2024graphneuralnetworks}.

\begin{algorithm}[b]
\caption{Graph $K$-oarsening (GK)}
\label{alg:koarsening}
\begin{algorithmic}[1]
\Require Step size $\lambda > 0$, ~~$K < N$, ~~$\delta > 0$, ~~$T \in \naturals$
\State Initialize $\bbTheta$ and $t=1$
\State Obtain $\bbP$ in~\eqref{eq:reduce} via $K$-means clustering of $f(\bbX,\bbA;\bbTheta)$
\State Initialize $\bbA' = \bbP^\top \bbA \bbP$, ~~
$\bbX' = \bbC^{-1}\bbP^\top \bbX$,
\Statex \qquad\qquad 
and $\bbZ_{\rm \scriptscriptstyle last} = f(\bbX, \bbA; \bbTheta)$
\While{not terminated}
    \State Obtain coarsened embeddings
        $\bbZ' \leftarrow f'(\bbX', \bbA'; \bbTheta)$
    \State Gradient update
        $\bbTheta \leftarrow \bbTheta - \lambda \nabla_{\bbTheta} \ccalL( \bby_{\rm \scriptscriptstyle train}, ~\, \bbP\bbZ' )$
    \State Update $t \leftarrow t + 1$
    \If{$\| \bbZ - \bbZ_{\rm \scriptscriptstyle last} \|_F / \| \bbZ_{\rm \scriptscriptstyle last} \|_F > \delta$ or $t > T$}
    \State Reset $t \leftarrow 1$
    \State Obtain node embeddings
        $\bbZ \leftarrow f(\bbX, \bbA; \bbTheta)$
    \State
        Update $\bbP$ in~\eqref{eq:reduce} via $K$-means clustering of $\bbZ$
    \State Update
        $\bbA' \leftarrow \bbP^\top \bbA \bbP$, ~~ $\bbX' \leftarrow \bbC^{-1}\bbP^\top \bbX$,
        ~~$\bbZ_{\rm \scriptscriptstyle last} \leftarrow \bbZ$
    \EndIf
\EndWhile
\State \Return $f(\cdot, \cdot; \bbTheta)$, $\bbP$
\end{algorithmic}
\end{algorithm}

\begin{table*}[t]
    \footnotesize
    \centering
    \caption{\small{Node classification accuracy on homophilic and heterophilic datasets. 
    The top performing methods at each ratio $r$ are {\bf bolded}.
    }}
    \vspace{.1cm}
    \begin{tabular}{c | c | c c c | c c c}
        \hline
        \multirow{2}{*}{\textbf{Dataset}}
            & \multirow{2}{*}{$r$}
            & \multicolumn{3}{ c| }{\textbf{AConvMatch}}
            & \multicolumn{3}{ c }{\textbf{Graph $K$-oarsening (GK)}} \\ 
            & 
            & \textbf{Val. acc.}
            & \textbf{Test acc.}
            & \textbf{Time (sec)}
            & \textbf{Val. acc.}
            & \textbf{Test acc.}
            & \textbf{Time (sec)} \\ \hline
            & $0.01$ \rule{0pt}{2ex}
            & {$\bf 0.65 \pm 0.00$} & {$\bf 0.64 \pm 0.00$} & {    $197.98 \pm 1.63$} 
            & {    $0.64 \pm 0.00$} & {    $0.63 \pm 0.00$} & {$\bf 129.21 \pm 1.32$} \\
        \textbf{OGBNArxiv}
            & $0.02$
            & {$0.66 \pm 0.00$} & {$0.65 \pm 0.00$} & {    $219.93 \pm 1.24$} 
            & {$0.66 \pm 0.00$} & {$0.65 \pm 0.00$} & {$\bf 188.21 \pm 0.61$} \\
        ($h=0.64$)
            & $0.05$
            & {$0.68 \pm 0.00$} & {    $0.67 \pm 0.00$} & {$\bf 420.38 \pm 3.80$} 
            & {$0.68 \pm 0.00$} & {$\bf 0.68 \pm 0.00$} & {    $427.47 \pm 2.67$} \\
            & $1.0$
            & {$\it 0.72 \pm 0.00$} & {$\it 0.71 \pm 0.00$} & {$\it 6855.59 \pm 119.78$} 
            & {$\it 0.72 \pm 0.00$} & {$\it 0.71 \pm 0.00$} & {$\it 6855.59 \pm 119.78$} 
            \\[.2ex] \hline
            & $0.05$ \rule{0pt}{2ex}
            & {$\bf 0.80 \pm 0.02$} & {$\bf 0.78 \pm 0.02$} & {    $52.99 \pm 0.54$} 
            & {    $0.76 \pm 0.02$} & {    $0.74 \pm 0.02$} & {$\bf 38.26 \pm 0.62$} \\
        \textbf{Cora}
            & $0.1$
            & {$\bf 0.81 \pm 0.01$} & {$\bf 0.80 \pm 0.02$} & {    $51.49 \pm 0.80$} 
            & {    $0.80 \pm 0.02$} & {    $0.77 \pm 0.02$} & {$\bf 39.61 \pm 0.39$} \\
        ($h=0.83$)
            & $0.25$
            & {$0.81 \pm 0.02$} & {    $0.79 \pm 0.02$} & {    $50.10 \pm 0.46$} 
            & {$0.81 \pm 0.02$} & {    $0.79 \pm 0.02$} & {$\bf 41.82 \pm 0.61$} \\
            & $1.0$
            & {$\it 0.81 \pm 0.02$} & {$\it 0.80 \pm 0.02$} & {$\it 35.84 \pm 1.11$} 
            & {$\it 0.81 \pm 0.02$} & {$\it 0.80 \pm 0.02$} & {$\it 35.84 \pm 1.11$} 
            \\[.2ex] \hline
            & $0.05$ \rule{0pt}{2ex}
            & {    $0.66 \pm 0.03$} & {    $0.65 \pm 0.04$} & {    $167.51 \pm 3.08$} 
            & {$\bf 0.70 \pm 0.03$} & {$\bf 0.68 \pm 0.02$} & {$\bf 47.66 \pm 1.45$} \\
        \textbf{Citeseer}
            & $0.1$
            & {    $0.67 \pm 0.04$} & {    $0.66 \pm 0.04$} & {    $163.13 \pm 3.20$} 
            & {$\bf 0.71 \pm 0.03$} & {$\bf 0.69 \pm 0.02$} & {$\bf 49.98 \pm 1.43$} \\
        ($h=0.71$)
            & $0.25$
            & {    $0.69 \pm 0.03$} & {    $0.68 \pm 0.03$} & {    $153.36 \pm 1.67$} 
            & {$\bf 0.72 \pm 0.03$} & {$\bf 0.70 \pm 0.02$} & {$\bf 56.92 \pm 1.41$} \\
            & $1.0$
            & {$\it 0.71 \pm 0.03$} & {$\it 0.69 \pm 0.02$} & {$\it 58.51 \pm 2.89$} 
            & {$\it 0.71 \pm 0.03$} & {$\it 0.69 \pm 0.02$} & {$\it 58.51 \pm 2.89$} 
            \\[.2ex] \hline
        \textbf{Wisconsin} \rule{0pt}{2ex}
            & $0.25$
            & {    $0.57 \pm 0.04$} & {    $0.52 \pm 0.06$} & {    $13.54 \pm 1.47$} 
            & {$\bf 0.74 \pm 0.03$} & {$\bf 0.70 \pm 0.08$} & {$\bf 9.22 \pm 0.62$} \\
        ($h=0.16$)
            & $1.0$
            & {$\it 0.76 \pm 0.04$} & {$\it 0.72 \pm 0.05$} & {$\it 7.63 \pm 0.30$} 
            & {$\it 0.76 \pm 0.04$} & {$\it 0.72 \pm 0.05$} & {$\it 7.63 \pm 0.30$} 
            \\[.2ex] \hline
        \textbf{Cornell} \rule{0pt}{2ex}
            & $0.25$
            & {    $0.64 \pm 0.08$} & {    $0.56 \pm 0.08$} & {    $11.81 \pm 1.59$} 
            & {$\bf 0.67 \pm 0.07$} & {$\bf 0.62 \pm 0.05$} & {$\bf 8.89 \pm 0.52$} \\
        ($h=0.11$)
            & $1.0$
            & {$\it 0.72 \pm 0.07$} & {$\it 0.64 \pm 0.06$} & {$\it 7.92 \pm 0.68$} 
            & {$\it 0.72 \pm 0.07$} & {$\it 0.64 \pm 0.06$} & {$\it 7.92 \pm 0.68$} 
            \\[.2ex] \hline
        \textbf{Texas} \rule{0pt}{2ex}
            & $0.25$
            & {    $0.64 \pm 0.04$} & {    $0.62 \pm 0.06$} & {    $11.24 \pm 1.17$} 
            & {$\bf 0.70 \pm 0.04$} & {$\bf 0.65 \pm 0.06$} & {$\bf 9.81 \pm 0.62$} \\
        ($h=0.06$)
            & $1.0$
            & {$\it 0.70 \pm 0.05$} & {$\it 0.70 \pm 0.06$} & {$\it 8.55 \pm 1.02$}
            & {$\it 0.70 \pm 0.05$} & {$\it 0.70 \pm 0.06$} & {$\it 8.55 \pm 1.02$}
        \\[.2ex] \hline
        \textbf{Chameleon} \rule{0pt}{2ex}
            & $0.1$
            & {    $0.44 \pm 0.02$} & {    $0.43 \pm 0.02$} & {    $66.73 \pm 3.37$} 
            & {$\bf 0.48 \pm 0.01$} & {$\bf 0.47 \pm 0.02$} & {$\bf 17.80 \pm 2.36$} \\
        ($h=0.25$)
            & $1.0$
            & {$\it 0.52 \pm 0.01$} & {$\it 0.50 \pm 0.02$} & {$\it 46.19 \pm 1.86$} 
            & {$\it 0.52 \pm 0.01$} & {$\it 0.50 \pm 0.02$} & {$\it 46.19 \pm 1.86$} 
        \\[.2ex] \hline
        \textbf{Squirrel} \rule{0pt}{2ex}
            & $0.1$
            & {    $0.29 \pm 0.02$} & {    $0.28 \pm 0.02$} & {    $232.68 \pm 4.61$} 
            & {$\bf 0.35 \pm 0.01$} & {$\bf 0.33 \pm 0.02$} & {$\bf 18.26 \pm 1.72$} \\
        ($h=0.22$)
            & $1.0$
            & {$\it 0.36 \pm 0.01$} & {$\it 0.35 \pm 0.01$} & {$\it 109.75 \pm 0.38$} 
            & {$\it 0.36 \pm 0.01$} & {$\it 0.35 \pm 0.01$} & {$\it 109.75 \pm 0.38$} 
        \\ \hline
    \end{tabular}\label{tab:results}
\end{table*}

\subsection{Related work}
\label{Ss:related}

A plethora of techniques exist to accelerate GNN training via graph reduction~\cite{xu2025LearningReduceScale,hashemi2024comprehensivesurveygraph}.
While scalable GNNs are more commonly explored via graph condensation~\cite{xu2024SurveyGraphCondensation,gao2025GraphCondensationSurvey,yin2025GraphCondensationFoundations}, we emphasize works that summarize graph data via coarsening~\cite{loukas2019GraphReductionSpectral,joly2025Taxonomyreductionmatrices}.
Previous methods merge nodes as a preprocessing step, where the coarsened graph is obtained prior to training, thus nodes are partitioned with no knowledge of model parameters~\cite{huang2021ScalingGraphNeural,dickens2024GraphCoarseningConvolution}.
Moreover, reducing a graph in preprocessing may not be amenable to scenarios with graphs that are noisy or experience changes in nodes or edges over time~\cite{hashemi2024comprehensivesurveygraph,gao2025RobGCrobustgraph}.
Furthermore, this also requires clustering nodes based solely on graph structure and node features.
Coarsening for GNNs has been designed to preserve graph properties or to be well suited to GCNs~\cite{huang2021ScalingGraphNeural,dickens2024GraphCoarseningConvolution,xia2025GraphCoarseningSupervised,joly2024Graphcoarseningmessagepassing}, but such techniques may struggle with settings such as heterophilic graph data, where edges are more likely to connect nodes of different classes~\cite{zheng2024graphneuralnetworks}.
As a relevant example, ConvMatch in~\cite{dickens2024GraphCoarseningConvolution} proposed hierarchical node clustering that merges nodes to solve
\alna{
    \min_{\bbP \in \ccalP}
    \left\| \theta \left(
        \bbP \tbA' \bbX' - \tbA \bbX
    \right) \right\|_{1,1}
\label{eq:convmatch}}
for $\tbA' := \tbD^{\prime-\frac{1}{2}} \bbP^\top (\bbA + \bbI) \bbP \tbD^{\prime-\frac{1}{2}}$ and $\tbD' := \diag( \bbP^\top (\bbA + \bbI) \bbP \bbone)$, where $\|\bbX\|_{1,1} = \sum_{i,j} |X_{ij}|$ and $\theta > 0$ denotes the parameter of a first-order graph filter to model the effect of a GCN. 
Thus, ConvMatch encourages similar \textit{filtered graph signals} between nodes in the original graph and their corresponding supernodes in the coarsened graph rather than promoting similar \textit{node embeddings} from the GNN $f$ of interest as in~\eqref{eq:lower_problem}.
While ConvMatch is effective for GCNs in~\eqref{eq:gcn}, coarsening via~\eqref{eq:convmatch} may not be suitable if we employ other architectures or if feature and edge variance within supernodes are relevant.
Differently, our approach can preserve information in $\bbA$ and $\bbX$ when coarsening via~\eqref{eq:lower_problem}.
Thus, we may find clusters suitable to the downstream task even if there is no natural partitioning of nodes based on the original structure or node features.


\section{Methodology}
\label{S:method}

We propose to solve~\eqref{eq:problem} via alternating minimization, shown in Algorithm~\ref{alg:koarsening}.
In particular, we first initialize GNN parameters $\bbTheta$, after which we perform $K$-means clustering on the output $\bbZ := f(\bbX,\bbA;\bbTheta)$ to solve the lower-level problem~\eqref{eq:lower_problem} given the initial $\bbTheta$~\cite{lloyd1982leastsquaresquantization}, yielding an initial coarsening $\bbA'$, $\bbX'$, and $\bbZ'$.
We then alternate between updating the GNN parameters $\bbTheta$ via gradient descent for the upper level problem~\eqref{eq:upper_problem} and solving the lower-level problem~\eqref{eq:lower_problem} to update the reduction matrix $\bbP$ via $K$-means clustering.
The gradient update in line 6 of Algorithm~\ref{alg:koarsening} lifts the coarsened embeddings to the original nodes to be used as predictions.
Thus, supernodes that yield more inaccurate predictions are given greater attention when updating $\bbTheta$.
We recoarsen $\bbP$ periodically every $T \in \naturals$ iterations to reduce training iteration complexity; however, to mitigate any error due to discrepancies in cluster assignments as embeddings drift, we also update $\bbP$ if node embeddings become sufficiently different as in line 7.
Additionally, since $K$-means does not guarantee convergence to optimal clusters, we initialize $\bbP$ in line 2 of Algorithm~\ref{alg:koarsening} by multiple instantiations of $K$-means with different starting conditions to increase the likelihood of encountering a desirable solution.
We further accelerate the recoarsening process in lines 7-10 by initializing $K$-means once, where we use the cluster assignments from the previous instance of coarsening.


We next highlight advantages of solving~\eqref{eq:problem} via Algorithm~\ref{alg:koarsening} relative to prior approaches.
Jointly obtaining GNN parameters $\bbTheta$ and the reduction matrix $\bbP$ via~\eqref{eq:problem} precludes the need for a potentially expensive preprocessing step, and, as we will show empirically, the additional coarsening steps in lines 7-10 of Algorithm~\ref{alg:koarsening} add little time to training while yielding benefits in terms of computation and performance.
Additionally, as mentioned in Section~\ref{Ss:related}, other coarsening approaches do not account for the GNN $f$, whereas we explicitly encourage similarities in embeddings, which is sufficient for matching GNN predictions or performance between the original and coarsened graphs~\cite{gao2025RethinkingAcceleratingGraph}.
However, unlike graph condensation methods that create synthetic graphs, we provide a direct, interpretable correspondence between original nodes and supernodes.
Thus, Algorithm~\ref{alg:koarsening} coarsens the original graph during training, allowing us to tailor clusters to the learning task without any preprocessing~\cite{xu2025LearningReduceScale}.

\begin{figure*}[t]
    \centering
    \begin{minipage}[b][][b]{0.2\textwidth}
        \begin{center}
            \includegraphics[width=\textwidth]{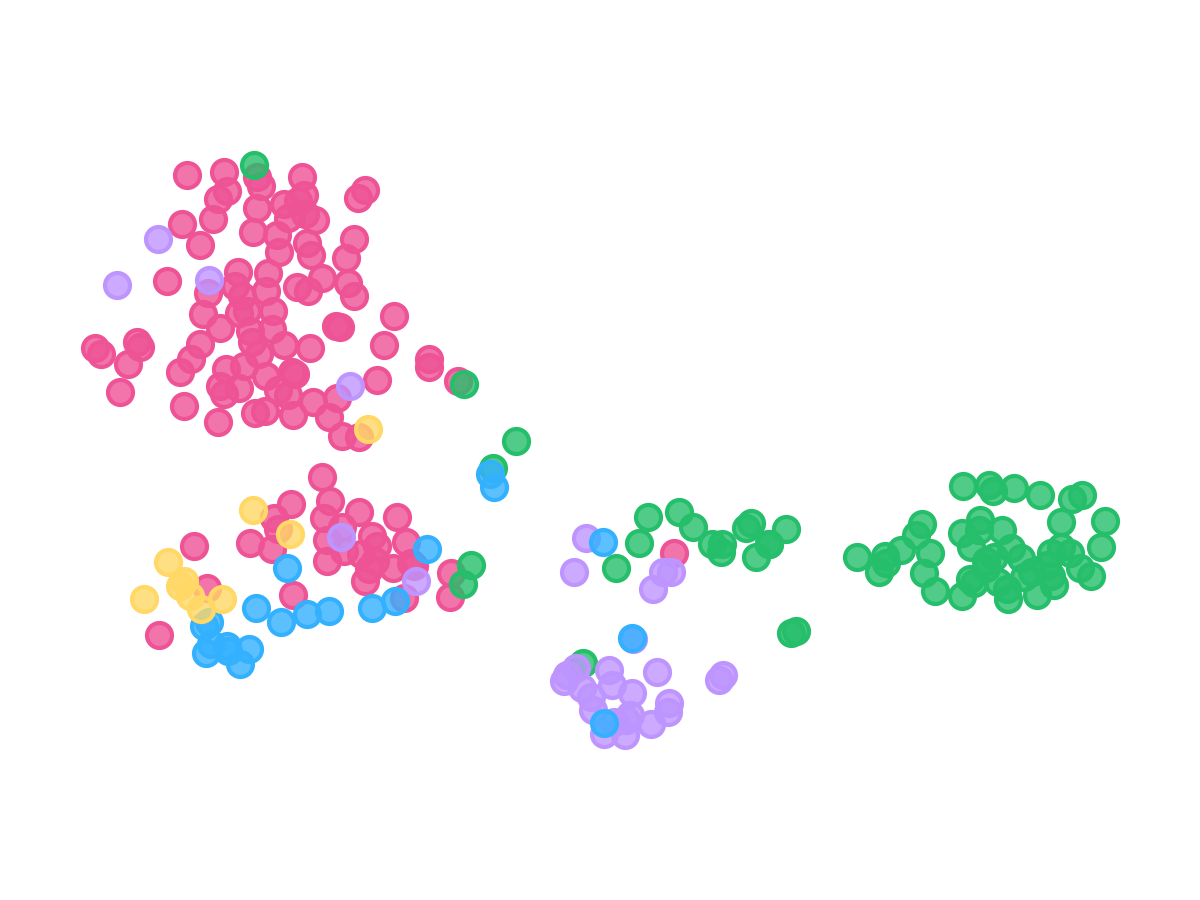}
            \small{(a)}
        \end{center}
    \end{minipage} 
    \hspace{.2cm}
    \begin{minipage}[b][][b]{0.2\textwidth}
        \begin{center}
            \includegraphics[width=\textwidth]{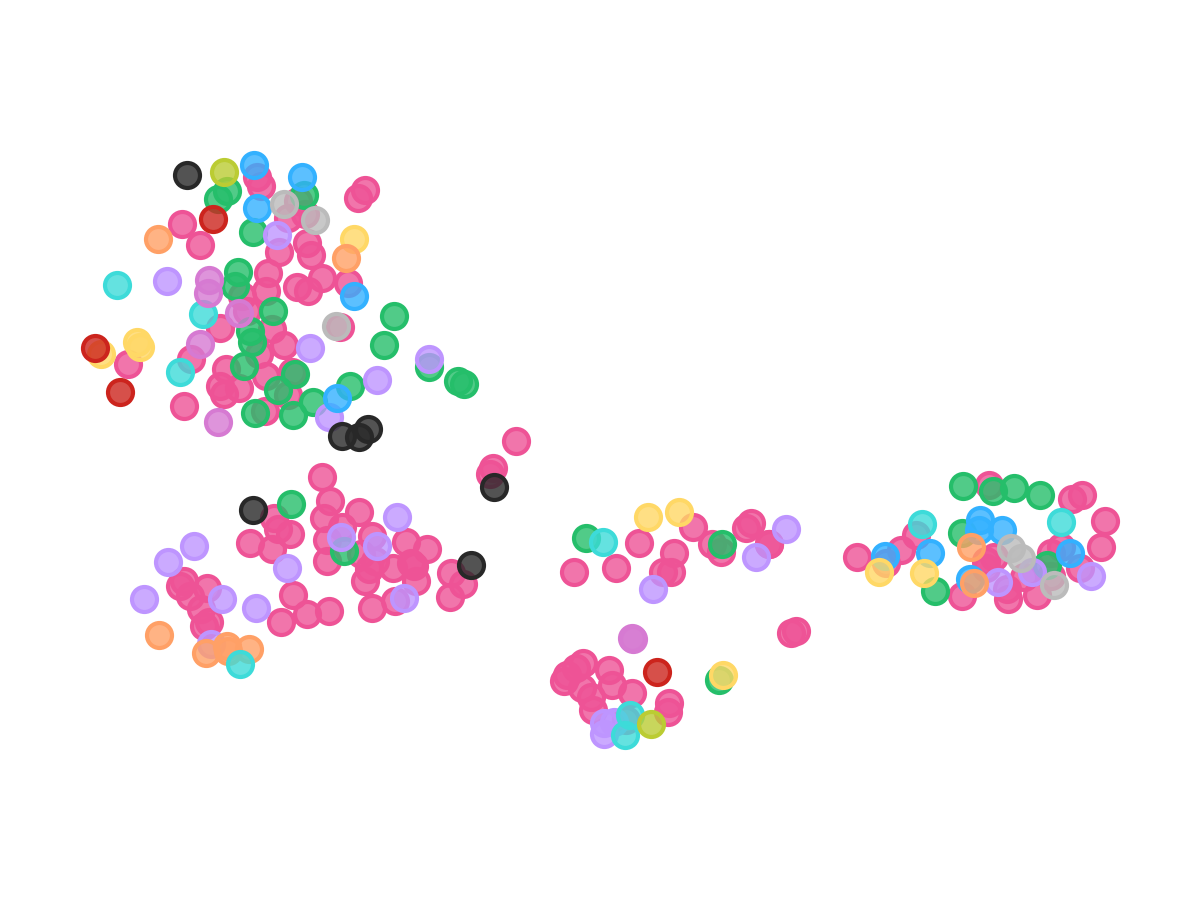}
            \small{(b)}
        \end{center}
    \end{minipage}
    \hspace{.2cm}
    \begin{minipage}[b][][b]{0.2\textwidth}
        \begin{center}
            \includegraphics[width=\textwidth]{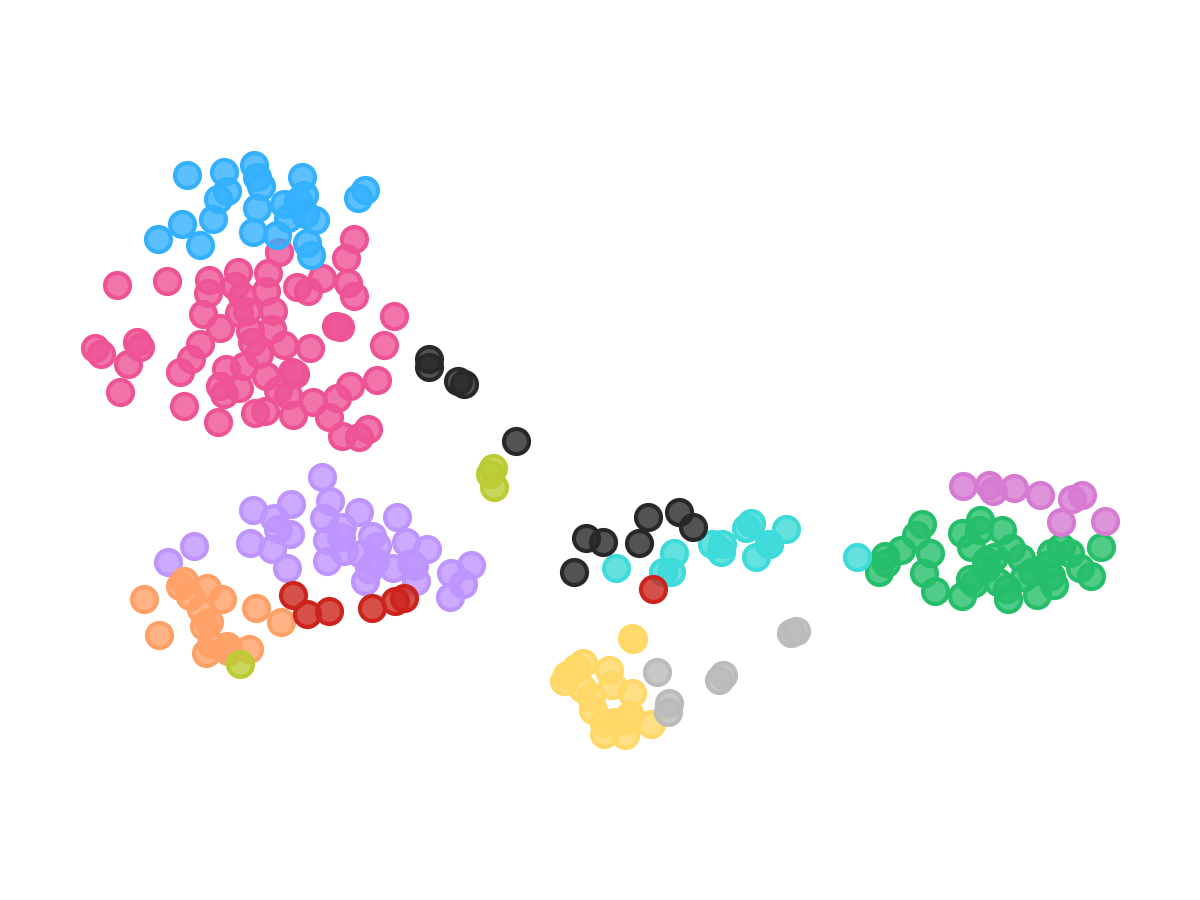}
            \small{(c)}
        \end{center}
    \end{minipage}
    \hspace{.2cm}
    \begin{minipage}[b][][b]{0.29\textwidth}
        \begin{center}
            \includegraphics[width=\textwidth]{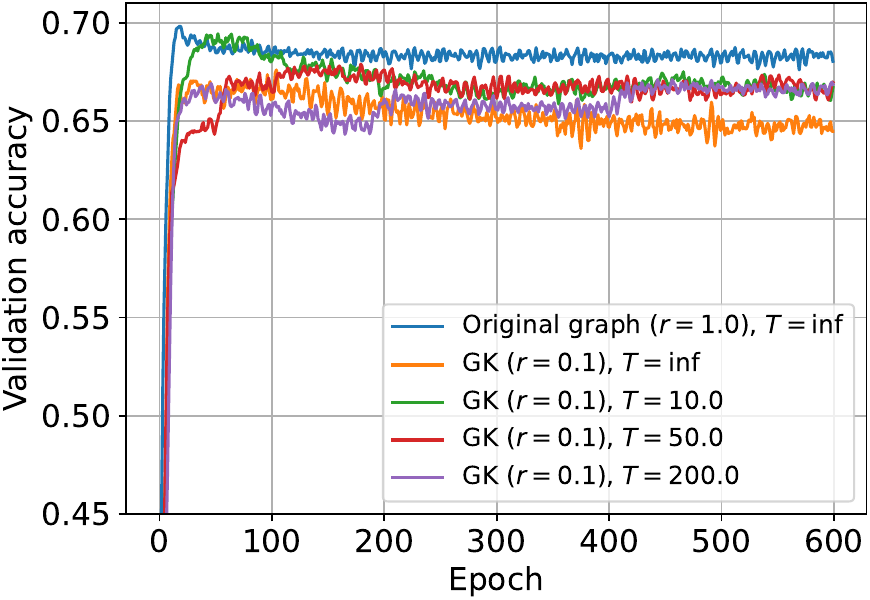}
            \small{(d)}
        \end{center}
    \end{minipage}
    \caption{
        (a)-(c) TSNE visualization of node embeddings for heterophilic \textbf{Wisconsin} dataset obtained from a model trained on the full, original graph ($r=1$).
        (a) Nodes colored by true class labels.
        (b) Nodes colored by \textbf{AConvMatch} cluster assignment.
        (c) Nodes colored by \textbf{GK} cluster assignment.
        (d) Validation trajectory as function of $T$ trained with \textbf{Citeseer} dataset ($r=1$) and with \textbf{GK} ($r=0.1$).
    }
    \label{f:embed}
\end{figure*}

\section{Numerical experiments}\label{S:experiments}

We evaluate our proposed approach solving~\eqref{eq:problem} with Algorithm~\ref{alg:koarsening} on the following benchmark datasets.
We consider both homophilic and heterophilic datasets to demonstrate the versatility of our approach in comparison with methods that assume a GCN-based model.
In particular, we compare our method \textbf{Graph $K$-oarsening (GK)} in Algorithm~\ref{alg:koarsening} to \textbf{AConvMatch} in~\cite{dickens2024GraphCoarseningConvolution}, which performs a faster approximation of~\eqref{eq:convmatch} while still yielding satisfactory results for homophilic graph data, outperforming multiple graph condensation and sparsification methods~\cite{sener2018active,jin22doscond}.
We evaluate both methods at multiple coarsening ratios $r := K / N$ for multiple datasets, and we implement a recoarsening period $T = 50$ for the following results. 
We consider three homophilic datasets 
(i) \textbf{OGBNArxiv} ($N=169343$)~\cite{hu2020ogb}, 
(ii) \textbf{Cora} ($N=2708$), 
and (iii) \textbf{Citeseer} ($N=3327$)~\cite{yang2016planetoid}, 
along with four heterophilic datasets~\cite{Pei2020Geom-GCN}
(iv) \textbf{Wisconsin} ($N=251$), 
(v) \textbf{Cornell} ($N=183$), 
(vi) \textbf{Texas} ($N=183$),
(vii) \textbf{Chameleon} ($N=2277$), and
(viii) \textbf{Squirrel} ($N=5201$).
Our results are shown in Table~\ref{tab:results}, which presents validation and testing accuracy for each method at every ratio $r$, along with the total training time.
For \textbf{AConvMatch}, we sum the coarsening and training durations, while \textbf{GK} only requires that we measure training time.
Further experimental details can be found at our GitHub repository\footnote{\href{https://github.com/RostyslavUA/graph_koarsening}{https://github.com/RostyslavUA/graph\_koarsening}}.

\subsection{Homophilic graph data}\label{Ss:homophilic}

We first compare node classification accuracy for training GCNs on coarsened graphs via \textbf{AConvMatch} and \textbf{GK} for the first three datasets in Table~\ref{tab:results}, along with performance when training a GCN on the full, original graph ($r=1$), highlighted in italics in Table~\ref{tab:results}.
We observe that \textbf{AConvMatch} and \textbf{GK} yield competitive accuracy for \textbf{OGBNArxiv}, \textbf{Cora}, and \textbf{Citeseer}.
However, \textbf{GK} tends to show greater improvement in computation time, particularly for Citeseer, for which we also see consistently superior accuracy, even relative to the original graph with a large enough ratio $r=0.25$.
While the full graph for the very sparse \textbf{Cora} trains the fastest, \textbf{GK} maintains training times more similar to the $r=1$ setting than \textbf{AConvMatch}.
For the larger \textbf{OGBNArxiv}, training a GCN on the full graph is unsurprisingly time consuming.
In comparison, both \textbf{AConvMatch} and \textbf{GK} improve total computation time, with both methods being competitive for a larger coarsened ratio $r = 0.05$.
However, despite maintaining similar performance to \textbf{AConvMatch}, \textbf{GK} shows noticeable improvements in training time for smaller graphs $r \in \{0.01, 0.02\}$, settings which are more critical for larger datasets.




\subsection{Heterophilic graph data}\label{Ss:heterophilic}

We next consider performance for heterophilic graph datasets, for which it is known that GCNs suffer in performance~\cite{zheng2024graphneuralnetworks}.
Thus, we instead train a FBGCN represented in~\eqref{eq:fbgcn} for each method in Table~\ref{tab:results}.
As before, we compare the graph coarsening methods to performance when using the full graph to train a FBGCN, denoted by $r=1$ for each dataset and italicized in Table~\ref{tab:results}.
Since \textbf{Wisconsin}, \textbf{Cornell}, and \textbf{Texas} are relatively small in size, we consider a ratio of $r=0.25$ to yield coarsened graphs that are small but still informative.
We also compare \textbf{AConvMatch} and \textbf{GK} using a larger heterophilic datasets \textbf{Chameleon} {and \textbf{Squirrel}, where we instead coarsen graphs with a ratio of $r = 0.1$.
As discussed previously, the method \textbf{AConvMatch} coarsens graphs assuming that data is homophilic.
Hence, for all five heterophilic datasets, \textbf{AConvMatch} exhibits a greater drop in accuracy compared to \textbf{GK}.
Indeed, our approach \textbf{GK} coarsens the graph based on node embeddings, so clusters are informed by graph data through the GNN without necessarily assuming homophilic data.
Thus, by choosing an appropriate architecture~\eqref{eq:fbgcn}, we observe consistently greater improvement in both training time and accuracy for \textbf{GK} on these heterophilic datasets.

To further investigate performance with heterophilic data, we visualize node embeddings and cluster assignments obtained via \textbf{AConvMatch} and \textbf{GK} in Fig.~\ref{f:embed}.
In each figure, we plot embeddings obtained from a FBGCN trained on the \textbf{Wisconsin} dataset using the full graph, where the left figure indicates the class labels of each node by color.
The node colors in the middle and right figures designate cluster assignments of nodes via \textbf{AConvMatch} and \textbf{GK}, respectively. 
We observe that \textbf{AConvMatch}, which coarsens graphs to preserve low-pass graph data~\cite{dickens2024GraphCoarseningConvolution}, clusters nodes belonging to different classes, as expected since adjacent nodes in the heterophilic \textbf{Wisconsin} graph are not likely to belong to the same class.
Thus, \textbf{AConvMatch} introduces a more challenging classification task, as the embeddings are not easily separable across classes.
In contrast, \textbf{GK} clusters nodes that mostly belong to the same class, indicating our ability to coarsen a graph based on its downstream task rather than relying on graph connectivity and node features alone.

Finally, to demonstrate the impact of the recoarsening period $T$ on training convergence, we plot the validation accuracy of \textbf{GK} during training for different values of $T$ in Fig.~\ref{f:embed}d.
We compare performance when training a GCN using the original \textbf{Citeseer} graph with $r=1$ versus a \textbf{GK} coarsened graph with $r=0.1$.
Compared to the setting where we perform $K$-means initially and do not recoarsen periodically $T = \infty$, we observe improved performance even with relatively large gaps between recoarsening $T = 200$.
As expected, reclustering more frequently increases accuracy, where \textbf{GK} is most competitive with the full graph setting when $T = 10$.
However, a smaller $T$ leads to a longer computation time, thus $T = 50$ strikes a balance between performance and efficiency for \textbf{Citeseer}.

\section{Conclusion}\label{S:conclusion}

In this work, we presented a graph coarsening approach that periodically clusters nodes based on their embeddings to adaptively accelerate GNN training.
As we demonstrated in our empirical results, we obtain embeddings that merge nodes based on relevance to the downstream task, including heterophilic data, which we found more challenging for methods that assume low-pass information.
Future work will see more advanced versions of the adaptive clustering steps, such as applying dynamic $K$-means clustering techniques to further accelerate the joint optimization process, rendering our approach amenable to continual learning~\cite{hashemi2024comprehensivesurveygraph}.
We also plan to demonstrate this work for other graph learning tasks, particularly unsupervised learning and graph-level predictions.

\bibliographystyle{IEEEbib}
\bibliography{refs}

\end{document}